\documentclass[sigconf,screen,nonacm]{acmart}

\makeatletter
\@ifundefined{Bbbk}{}{}
\makeatother

\usepackage{graphicx}
\usepackage{booktabs}
\usepackage{xcolor}
\usepackage{multirow}

\graphicspath{{./}}

\renewcommand\footnotetextcopyrightpermission[1]{}

\newcommand{\MaybeIncludeGraphic}[2][]{%
  \IfFileExists{#2}{\includegraphics[#1]{#2}}{%
    \fbox{\parbox[c][0.23\textheight][c]{0.95\linewidth}{\centering
      \texttt{Missing figure file:}\\\texttt{#2}}}%
  }%
}

\DeclareMathOperator{\Normalize}{norm}

\newcommand{\best}[1]{\textbf{#1}}
\newcommand{\na}{--}

\newcommand{\tblcommon}{%
  \small
  \renewcommand{\arraystretch}{1.12}%
  \setlength{\tabcolsep}{2.4pt}%
}

\newcommand{\tblwide}{%
  \small
  \renewcommand{\arraystretch}{1.08}%
  \setlength{\tabcolsep}{2.0pt}%
}

\title[CoDA]{CoDA: Exploring Chain-of-Distribution Attacks and Post-Hoc Token-Space Repair for Medical Vision-Language Models}

\author{Xiang Chen, Fangfang Yang, Chunlei Meng, Yuxian Dong, Ang Li, Yiwei Wei, Jiahuan Long, Jiujiang Guo, Chengyin Hu}

\begin{document}

% ------------------------------------------------------------
% Abstract
% ------------------------------------------------------------

\begin{abstract}
Medical vision-language models (MVLMs) are increasingly used as perceptual backbones in radiology pipelines and as the visual front end of multimodal assistants, yet their reliability under real clinical imaging workflows remains underexplored. Prior robustness evaluations often assume clean, curated inputs or study isolated corruptions and pixel-level perturbations, overlooking routine acquisition, reconstruction, display, and delivery operations that preserve clinical readability while subtly shifting image statistics. To address this gap, we propose CoDA, a chain-of-distribution framework that constructs clinically plausible pipeline shifts by composing acquisition-like shading, reconstruction and display remapping, and delivery and export degradations. Under masked structural-similarity constraints, CoDA jointly optimizes stage-composition families and within-stage parameters to induce failures while preserving visual plausibility. Across brain MRI, chest X-ray, and abdominal CT, CoDA substantially degrades the zero-shot performance of CLIP-style MVLMs, with ablations showing that chained compositions are consistently more damaging than any single stage. We also evaluate multimodal large language models (MLLMs) as technical-authenticity auditors of imaging realism and quality rather than pathology. Proprietary MLLMs show degraded auditing reliability and persistent high-confidence errors on CoDA-shifted samples under a single-call self-challenge protocol, while the medical-specific MLLMs we test exhibit clear deficiencies in medical image quality auditing. Motivated by these findings, we introduce a post-hoc repair strategy based on teacher-guided token-space adaptation with patch-level alignment to improve zero-shot classification accuracy on archived CoDA outputs. Overall, our study characterizes a clinically grounded threat surface for MVLM deployment and shows that lightweight alignment improves robustness in deployment.
\end{abstract}

% ------------------------------------------------------------
% Keywords
% ------------------------------------------------------------
\keywords{Medical vision-language models, Robustness and security evaluations, CoDA, Post-hoc repair strategy}

% ============================================================
% Figure 1 only (Teaser)
% ============================================================
\begin{teaserfigure}
  \centering
  \includegraphics[width=\textwidth]{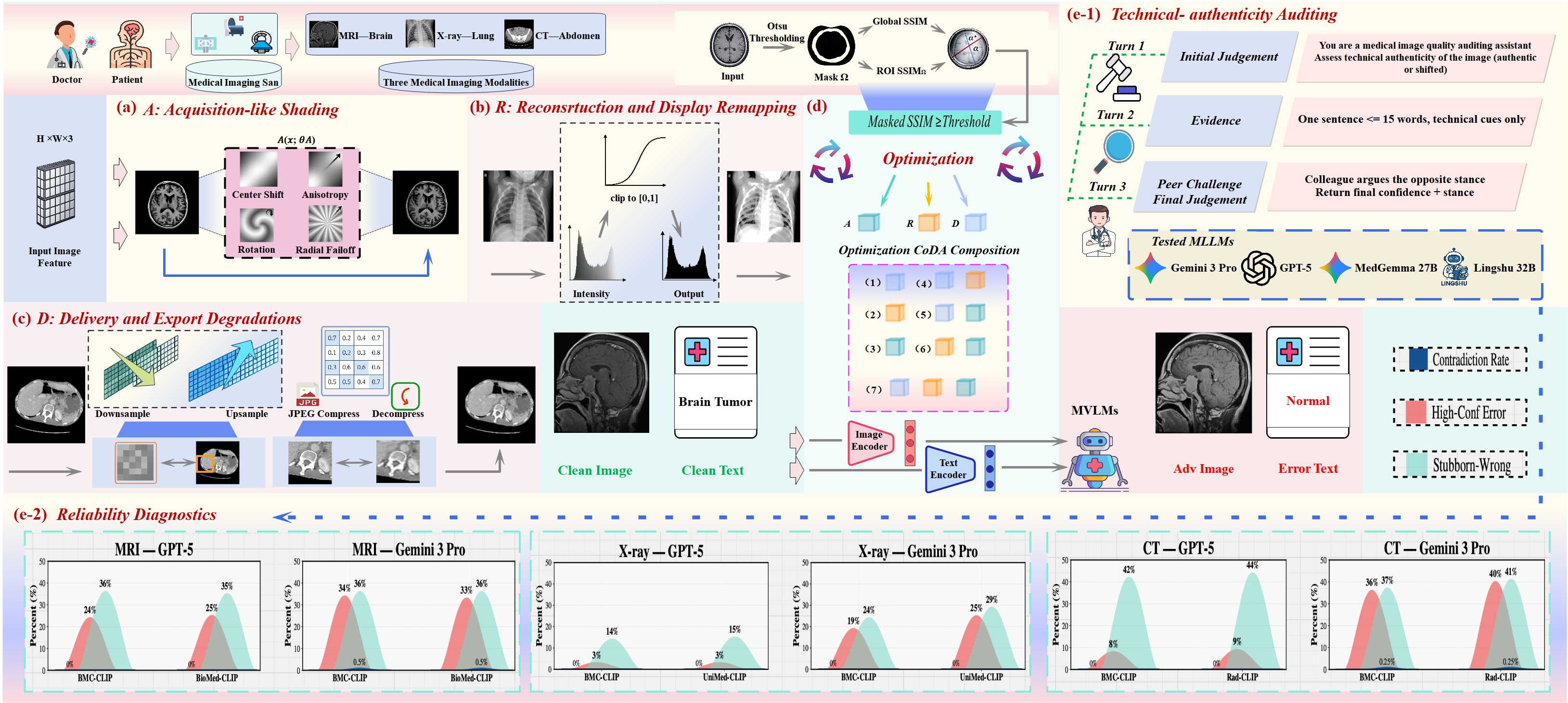}
  \caption{\textbf{Overview of CoDA.}
  \textbf{(a)\textasciitilde(c) Three pipeline stages: acquisition-like shading (\boldmath$A$\unboldmath), reconstruction and display remapping (\boldmath$R$\unboldmath), and delivery and export degradations (\boldmath$D$\unboldmath), illustrated on brain MRI, chest X-ray, and abdominal CT.}
  \textbf{(d) CoDA optimizes over seven stage compositions and within-stage parameters under masked similarity constraints to produce visually plausible adversarial samples.}
  \textbf{(e-1)\textasciitilde(e-2) Technical-authenticity auditing and reliability diagnostics.}}
  \Description{An overview figure of CoDA. Panels (a) to (c) illustrate three clinically plausible pipeline stages on brain MRI, chest X-ray, and abdominal CT: acquisition-like shading, reconstruction and display remapping, and delivery and export degradations. Panel (d) shows that CoDA searches over seven stage-composition families and their within-stage parameters under masked similarity constraints to generate visually plausible adversarial samples. Panels (e-1) to (e-2) present the technical-authenticity auditing setup and related reliability diagnostics.}
  \label{fig:overview}
\end{teaserfigure}

\maketitle

% ============================================================
% 1 INTRODUCTION
% ============================================================
\section{Introduction}

MVLMs are increasingly adopted as general-purpose perceptual backbones in radiology AI, enabling zero-shot recognition and retrieval, report-aware representation learning, and language-guided visual grounding. Their progress has been driven by contrastive pretraining and domain-specific adaptation, including CLIP-style vision--language pretraining~\cite{radford2021clip}, medical image--text contrastive learning~\cite{zhang2022convirt}, report-based radiology representation learning~\cite{zhou2022refers}, and large-scale biomedical figure--caption pretraining~\cite{zhang2023biomedclip}. Specialized CLIP variants for radiology and expanding biomedical archives have further improved medical encoders~\cite{lu2024radclip,biomedica2025}, while visually instructed MLLMs are emerging in parallel as multimodal assistants~\cite{liu2023llava,gemini2023,openai2023gpt4}. Since the vision encoder governs visual evidence for downstream reasoning, front-end robustness is safety-critical.

Clinical images, however, rarely arrive as pristine tensors. They are shaped by imaging workflows in which acquisition conditions affect intensity profiles, reconstruction and display settings remap physical measurements to pixels (e.g., window and level in DICOM viewers)~\cite{dicom2024}, and delivery operations such as resampling and compression introduce systematic artifacts~\cite{bourai2024compression}. These steps usually preserve human interpretability, yet they can alter low-level statistics in ways that MVLM representations may depend on. More broadly, distribution shift is known to degrade deployed models~\cite{zech2018pneumonia,degrave2021covidshortcut,varma2024ravl,noohdani2024dac,huang2025framesvqa}, and recent robustness benchmarks for large vision--language models show that routine, non-malicious perturbations can induce substantial failures even when semantics remain intact~\cite{tu2025unicorn,zhao2023attackvlm,schlarmann2024robustclip,xie2025chainofattack,bhagwatkar2024improving,bhagwatkar2024towardsrobust}. Nevertheless, robustness evaluations for MVLMs still often emphasize clean public datasets or isolated perturbations, leaving limited coverage of multi-stage variability introduced by real clinical pipelines.

We address this gap with an adversarial yet clinically grounded threat model. Rather than considering classic norm-bounded pixel attacks, we study an attacker who manipulates plausible pipeline operations and their parameters through realistic acquisition, reconstruction, and delivery processes. We propose CoDA, a chain-of-distribution attack that composes three stages: acquisition-like shading, reconstruction and display remapping, and delivery and export degradations. Under masked structural-similarity constraints that preserve visual plausibility~\cite{wang2004ssim}, CoDA optimizes failure-inducing parameterizations over both composition families and within-stage parameters. This design reflects clinical workflows in which deviations accumulate across multiple processing steps and interact under \mbox{realistic clinical operating conditions}.

Using CoDA, we show that chained pipeline shifts substantially undermine zero-shot CLIP-style MVLMs across brain MRI, chest X-ray, and abdominal CT. Ablations restricting the shift space to individual stages confirm that chained compositions are consistently more damaging than any single-stage variant under the same plausibility constraints. We then examine whether MLLM-based multimodal assistants can serve as a safety layer by auditing technical authenticity, assessing imaging realism and quality rather than pathology. Motivated by evidence that MLLMs can hallucinate visual content and remain confidently wrong~\cite{li2023pope,guan2024hallusionbench}, we adapt reliability diagnostics to the medical pipeline setting and evaluate these auditors under CoDA-shifted inputs. This setting is particularly challenging because the induced shifts remain visually plausible while perturbing the technical cues that auditing-oriented systems are expected to detect. We observe that strong proprietary MLLMs remain reliable on clean inputs but fail to consistently flag adversarial pipeline outputs, even under self-challenge-style prompting~\cite{wei2022cot,wang2022selfconsistency}. More strikingly, the evaluated medical-specific MLLMs exhibit clear deficiencies in medical image quality auditing, limiting their utility as a quality-control layer under realistic clinical pipeline variability in practice.

We further investigate a lightweight post-hoc repair strategy to improve robustness without full retraining. The design combines parameter-efficient adaptation methods such as LoRA and SSF~\cite{hu2022lora,lian2022ssf} with ideas from knowledge distillation~\cite{gou2021kdsurvey}, efficient visual adaptation~\cite{he2023sensitivityaware,mercea2024timememory}, and recent robustness-oriented adaptation for vision--language models~\cite{liu2026adpo,choi2026advmask}. Specifically, we introduce a teacher-guided token-space adaptation scheme that aligns a student vision encoder with the token structure of a stronger teacher while preserving the student's interface and computational profile. Evaluated on archived CoDA outputs, this repair improves MVLM zero-shot classification accuracy, suggesting that post-hoc alignment can recover performance under clinically plausible deployment-time pipeline shifts. We summarize our main contributions as follows:

\begin{itemize}
  \item We propose CoDA, a clinically grounded attack space that composes acquisition-like shading, reconstruction and display remapping, and delivery and export degradations, and optimizes both the stage composition and the stage parameters under masked similarity constraints.
  \item We conduct a multi-stage robustness evaluation of CLIP-style MVLMs on brain MRI, chest X-ray, and abdominal CT, including controlled ablations that isolate each stage and budget analyses that quantify how attack strength scales with optimization iterations.
  \item We show that CoDA degrades zero-shot MVLM performance and weakens MLLM auditing of technical authenticity under pipeline shifts. We further demonstrate that lightweight teacher-guided token-space repair can substantially recover robustness without full retraining, while remaining compatible with practical deployment requirements in routine clinical imaging workflows.
\end{itemize}

% ============================================================
% Table 2
% ============================================================
\begin{table*}[!t]
\centering
\tblwide
\caption{\textbf{Attack success rate (\%).} Each entry gives the success rate after optimizing the within-family parameters for its CoDA composition family. \textbf{Bold} indicates the best within each modality block (MRI/X-ray/CT) for the same model and setting.}
\label{tab:family_success}
\begin{tabular*}{\textwidth}{@{\extracolsep{\fill}}ll*{7}{c}*{7}{c}*{7}{c}@{}}
\toprule
\multirow{2}{*}{Model} & \multirow{2}{*}{Setting}
& \multicolumn{7}{c}{MRI}
& \multicolumn{7}{c}{X-ray}
& \multicolumn{7}{c}{CT} \\
\cmidrule(lr){3-9}\cmidrule(lr){10-16}\cmidrule(lr){17-23}
& & $A$ & $R$ & $D$ & $A{+}R$ & $R{+}D$ & $A{+}D$ & $A{+}R{+}D$
  & $A$ & $R$ & $D$ & $A{+}R$ & $R{+}D$ & $A{+}D$ & $A{+}R{+}D$
  & $A$ & $R$ & $D$ & $A{+}R$ & $R{+}D$ & $A{+}D$ & $A{+}R{+}D$ \\
\midrule

\multirow{4}{*}{BioMed-CLIP}
& Random ($\tau=0.90$)
& 11 & 40 & 33 & 35 & \best{54} & 26 & 38
& 71 & 61 & \best{75} & 33 & 45 & \best{75} & 24
& 69 & 73 & 67 & 66 & 68 & 67 & \best{74} \\
& Random ($\tau=0.80$)
& 22 & 61 & 17 & 52 & \best{62} & 25 & 44
& 75 & 77 & \best{100} & 53 & 63 & 67 & 61
& 69 & \best{88} & 50 & 70 & 81 & 75 & 76 \\
& Bayes ($\tau=0.90$)
& 0 & \best{61} & 25 & 45 & 58 & 7 & 45
& 69 & 68 & 90 & 44 & 42 & \best{100} & 37
& 62 & 76 & 60 & 64 & 68 & \best{87} & 79 \\
& Bayes ($\tau=0.80$)
& 0 & 51 & \best{100} & 61 & 47 & 45 & 68
& 73 & 73 & 90 & 70 & 85 & \best{100} & 78
& 71 & 76 & 54 & 70 & 92 & \best{94} & 79 \\
\midrule

\multirow{4}{*}{UniMed-CLIP}
& Random ($\tau=0.90$)
& \best{83} & 21 & 71 & 19 & 48 & 80 & 50
& 64 & 50 & 90 & 55 & \best{100} & 70 & 55
& 76 & 86 & 67 & 84 & 76 & 79 & \best{90} \\
& Random ($\tau=0.80$)
& 67 & 26 & 69 & 37 & 47 & \best{83} & 59
& 40 & 33 & 85 & 30 & \best{100} & 60 & 80
& 89 & 87 & 81 & 90 & \best{97} & 92 & 71 \\
& Bayes ($\tau=0.90$)
& 75 & 31 & \best{79} & 23 & 35 & 66 & 68
& 0 & 58 & \best{95} & 37 & 90 & 4 & 75
& 80 & \best{92} & 68 & 91 & 89 & 88 & 86 \\
& Bayes ($\tau=0.80$)
& 67 & 27 & 70 & 41 & 57 & 73 & \best{74}
& 35 & 45 & \best{97} & 30 & 88 & 76 & 64
& 90 & 91 & 88 & 91 & \best{97} & 91 & 89 \\
\midrule

\multirow{4}{*}{BMC-CLIP}
& Random ($\tau=0.90$)
& 0 & 42 & 47 & 33 & \best{54} & 20 & 42
& 0 & 19 & 96 & 6 & \best{100} & \best{100} & 50
& 80 & \best{83} & 29 & 77 & 56 & 21 & 50 \\
& Random ($\tau=0.80$)
& 0 & 47 & 45 & 42 & \best{49} & 47 & 38
& 0 & 23 & 97 & 15 & \best{100} & 91 & 40
& 78 & \best{83} & 30 & 72 & 59 & 18 & 45 \\
& Bayes ($\tau=0.90$)
& 0 & 30 & 47 & 33 & \best{50} & 40 & 48
& 0 & 33 & 95 & 11 & 94 & \best{100} & 75
& 83 & \best{84} & 37 & 69 & 48 & 23 & 37 \\
& Bayes ($\tau=0.80$)
& 0 & 39 & 47 & 33 & \best{60} & 27 & 52
& 9 & 29 & \best{97} & 22 & 95 & 80 & 64
& 83 & \best{90} & 21 & 66 & 45 & 33 & 49 \\
\midrule

\multirow{4}{*}{Rad-CLIP}
& Random ($\tau=0.90$)
& 73 & 16 & \best{100} & 20 & 27 & 33 & 15
& \na & 50 & 50 & \best{85} & 36 & 20 & 67
& \best{100} & 45 & 32 & 69 & 42 & 53 & 71 \\
& Random ($\tau=0.80$)
& 58 & 29 & \best{100} & 48 & 36 & 79 & 45
& \na & 58 & 50 & \best{88} & 57 & 43 & 81
& \best{94} & 88 & 61 & 80 & 45 & 61 & 58 \\
& Bayes ($\tau=0.90$)
& 56 & 15 & 50 & 34 & 27 & \best{62} & 23
& 0 & 52 & 50 & \best{87} & 33 & 25 & 77
& \best{82} & 57 & 38 & 66 & 49 & 53 & 62 \\
& Bayes ($\tau=0.80$)
& \best{67} & 38 & \na & 50 & 50 & 64 & 59
& \na & 61 & 71 & \best{91} & 44 & 0 & 85
& \best{100} & 90 & 50 & 82 & 55 & 69 & 69 \\
\bottomrule
\end{tabular*}
\end{table*}

% ============================================================
% 2 RELATED WORK
% ============================================================
\section{Related Work}

\subsection{Robustness and Distribution Shift}
Recent work on robustness increasingly emphasizes structured perturbations and realistic distribution shifts, rather than synthetic worst-case noise alone. In multimodal and vision--language settings, robustness degrades under adversarial attacks~\cite{zhao2023attackvlm,schlarmann2024robustclip,xie2025chainofattack}, as well as under compositional shifts, spurious correlations, and changing evaluation settings~\cite{varma2024ravl,bhagwatkar2024improving,bhagwatkar2024towardsrobust,huang2025framesvqa,tu2025unicorn}. This perspective is especially relevant for real-world clinical deployment, where images may undergo acquisition variability, reconstruction changes, export artifacts, and other workflow-induced transformations while remaining clinically interpretable and diagnostically usable. Our work follows this line by targeting clinically plausible multi-stage pipeline variability across acquisition, reconstruction, and delivery processes in realistic medical imaging workflows.

\subsection{Medical Vision--Language Pretraining}
MVLMs extend CLIP-style contrastive learning by leveraging paired reports, biomedical captions, and large-scale medical corpora to acquire transferable representations for zero-shot transfer and cross-task generalization in radiology~\cite{zhang2022convirt,zhou2022refers,zhang2023biomedclip,lu2024radclip,biomedica2025}. These advances have substantially broadened the practical scope of medical vision--language pretraining across recognition and retrieval tasks in diverse radiological settings and institutions worldwide. However, deployment reliability remains strongly shaped by heterogeneity in scanners, acquisition protocols, reconstruction settings, and downstream processing, where small yet systematic variations can alter pixel statistics without compromising clinical interpretability. Prior medical AI studies have shown that dataset bias and shortcut features can inflate benchmark performance while degrading generalization under real-world distribution shift~\cite{zech2018pneumonia,degrave2021covidshortcut}. Complementing this line of work, we study pipeline-consistent shift spaces that stress-test MVLMs under clinically plausible variability and show how compounded stages can expose vulnerabilities often missed by standard single-stage evaluations in prior robustness studies of medical imaging systems under realistic deployment conditions.

\subsection{Multimodal Reliability and Security}
MLLMs expand the threat surface beyond pixel robustness to the broader image--language interface. In medical settings, reliability can be compromised by prompt injection~\cite{clusmann2025promptinjection}, jailbreak and cross-modality mismatch behaviors~\cite{huang2025medmllmvulnerable}, backdoor vulnerabilities in medical image--text models~\cite{jin2024medclipbackdoor}, and model stealing under adversarial domain alignment~\cite{shen2025adasteal}. These studies indicate that safe clinical deployment depends on the stability of the full image--language interface, not visual robustness alone. A related but less explored question is whether MLLMs can act as robust technical safeguards by auditing image realism and acquisition quality. Our results further show that this capability is itself fragile under clinically plausible chained shifts: proprietary MLLMs degrade on CoDA outputs, while the evaluated medical-specific MLLMs exhibit limited and unreliable capability for practical medical image quality auditing under these realistic and highly challenging deployment conditions.

% ============================================================
% Table 1
% ============================================================
\begin{table*}[!t]
 \caption{\textbf{Zero-shot accuracy (\%).} Zero-shot classification performance of CLIP-style MVLMs on clean inputs and under CoDA shifts across MRI, X-ray, and CT, evaluated with random and Bayesian optimization at two plausibility thresholds.}
  \label{tab:zscls}
  \centering
  \tblcommon
  \begin{tabular*}{\textwidth}{@{\extracolsep{\fill}}lccc ccc ccc ccc@{}}
    \toprule
    \multirow{2}{*}{Setting} &
    \multicolumn{3}{c}{BioMed-CLIP} &
    \multicolumn{3}{c}{UniMed-CLIP} &
    \multicolumn{3}{c}{BMC-CLIP} &
    \multicolumn{3}{c}{Rad-CLIP} \\
    \cmidrule(lr){2-4}\cmidrule(lr){5-7}\cmidrule(lr){8-10}\cmidrule(lr){11-13}
     & MRI & X-ray & CT & MRI & X-ray & CT & MRI & X-ray & CT & MRI & X-ray & CT \\
    \midrule
    Clean & 81.5 & 64.0 & 57.5 & 65.0 & 58.5 & 54.0 & 78.5 & 50.5 & 54.0 & 93.0 & 85.5 & 50.0 \\
    $\tau=0.9$ / Random & 62.5 & 52.5 & 31.0 & 51.5 & 51.0 & 21.0 & 62.0 & 40.5 & 44.0 & 76.0 & 37.0 & 44.5 \\
    $\tau=0.8$ / Random & 50.5 & 32.0 & 25.0 & 47.5 & 46.5 & 11.5 & 57.5 & 38.0 & 42.5 & 56.5 & 31.5 & 33.0 \\
    $\tau=0.9$ / Bayes & 56.5 & 45.5 & 29.0 & 48.0 & 48.5 & 15.5 & 59.5 & 38.0 & 44.5 & 71.5 & 33.5 & 44.0 \\
    $\tau=0.8$ / Bayes & 44.0 & 24.5 & 22.0 & 44.0 & 42.5 & 8.5 & 57.0 & 32.5 & 40.0 & 50.5 & 28.0 & 30.0 \\
    \bottomrule
  \end{tabular*}
\end{table*}

% ============================================================
% 3 METHODOLOGY
% ============================================================
\section{Methodology}

\newcommand{\paren}[1]{\bigl(#1\bigr)}

\subsection{Problem Definition}
We study CLIP-style MVLMs consisting of an image encoder $f_{\mathit{img}}$ and a text encoder $f_{\mathit{text}}$~\cite{radford2021clip}. Given an image $x\in\mathbb{R}^{H\times W\times 3}$, we use the $\ell_2$-normalized image embedding:
\begin{equation}
\mathbf{z}(x)=\frac{f_{\mathit{img}}(x)}{\|f_{\mathit{img}}(x)\|_2}\in\mathbb{R}^{d}
\label{eq:z}
\end{equation}
For zero-shot classification evaluation, we follow prompt ensembling~\cite{radford2021clip}. For each class $c\in\{0,1\}$ with prompt set $\mathcal{T}_c$, we construct the corresponding class prototype representation:
\begin{equation}
\mathbf{t}_c=\frac{1}{|\mathcal{T}_c|}\sum_{t\in\mathcal{T}_c}\frac{f_{\mathit{text}}(t)}{\|f_{\mathit{text}}(t)\|_2}\in\mathbb{R}^{d}
\label{eq:tc}
\end{equation}
We score classes by cosine similarity $s_c(x)=\langle \mathbf{z}(x),\mathbf{t}_c\rangle$ and define the margin as $m(x)=s_1(x)-s_0(x)$. Given label $y\in\{0,1\}$, we use a signed correctness score:
\begin{equation}
J(x,y)=m(x)\cdot\sigma(y)\qquad
\sigma(y)=\begin{cases}
+1,& y=1\\
-1,& y=0
\end{cases}
\label{eq:J}
\end{equation}
where $\sigma(\cdot)$ maps the binary label to $\pm1$, so that $J(x,y)<0$ indicates a prediction flip.

\subsection{CoDA Shift Space and Compositions}
CoDA models pipeline variability using three parameterized stages: acquisition-like shading ($A$), reconstruction and display remapping ($R$), and delivery and export degradations ($D$). For the transformations below, we assume an intensity-normalized input image, denote $x \in [0,1]^{H\times W\times 3}$, and use $\operatorname{clip}(\cdot)$ to clamp values to $[0,1]$.

\noindent\textbf{Stage \boldmath$A$\unboldmath\ (Acquisition-like Shading).}
We apply a smooth multiplicative gain field $M_A(\cdot;\theta_A)\in\mathbb{R}^{H\times W}$ to model acquisition-induced, spatially varying intensity gains (e.g., center shift, anisotropy, rotation, and radial falloff) in routine clinical imaging. Fig.~\ref{fig:overview} (a) shows representative examples of this effect:
\begin{equation}
A(x;\theta_A)=\operatorname{clip}\bigl(x\odot M_A(x;\theta_A)\bigr)
\label{eq:A_compact}
\end{equation}
Here, $\odot$ denotes element-wise multiplication broadcast over channels, and $\theta_A$ collects the gain strength and geometric parameters that define $M_A$.

\noindent\textbf{Stage \boldmath$R$\unboldmath\ (Reconstruction and Display Remapping).}
To capture reconstruction and viewing variability, we apply robust window--level normalization followed by a monotone tone curve, gamma adjustment, and bit-depth quantization. These steps mirror common clinical viewing and conversion toolchains~\cite{dicom2024}, with qualitative illustrations provided in Fig.~\ref{fig:overview} (b):
\begin{equation}
R(x;\theta_R)
=Q_b\Bigl(
T\Bigl(
\operatorname{clip}\Bigl(\frac{x-c}{w+\varepsilon}+\frac{1}{2}\Bigr)
\Bigr)^{\gamma}
\Bigr)
\label{eq:R_compact}
\end{equation}
Here, $(c,w)$ are the window center and width computed from robust quantiles of a grayscale proxy and then adjusted by offsets $(\Delta_c,s_w)$, $\varepsilon$ is a small constant for numerical stability, $T(\cdot)$ is a piecewise-linear monotone tone curve defined by control points $(0,0)$, $(0.25,y_{25})$, $(0.5,y_{50})$, $(0.75,y_{75})$, and $(1,1)$ with monotonicity enforced, and $Q_b(\cdot)$ denotes $b$-bit uniform quantization. We denote $\theta_R=\{\Delta_c,\, s_w,\, y_{25},\, y_{50},\, y_{75},\, \gamma,\, b\}$.

\noindent\textbf{Stage \boldmath$D$\unboldmath\ (Delivery and Export Degradations).}
We model delivery-time artifacts using downsample--upsample resizing $\mathcal{S}_{\rho}(\cdot)$ and compression-induced degradation $\mathcal{J}_q(\cdot)$, which together approximate common export and transmission artifacts in clinical image pipelines~\cite{bourai2024compression}. The resulting degradations are shown in Fig.~\ref{fig:overview} (c):
\begin{equation}
D(x;\theta_D)=\mathcal{J}_q\bigl(\mathcal{S}_{\rho}(x)\bigr)
\label{eq:D_compact}
\end{equation}
Here, $\rho\in(0,1]$ is the resize factor, $q$ is the compression-quality factor, and $\theta_D=\{q,\,\rho\}$.

\noindent\textbf{Compositions.}
CoDA evaluates seven composition families under the canonical stage order $A\rightarrow R\rightarrow D$:
\begin{equation}
\mathcal{F}=\{A,\,R,\,D,\,A\!\circ\!R,\,R\!\circ\!D,\,A\!\circ\!D,\,A\!\circ\!R\!\circ\!D\}
\label{eq:families_compact}
\end{equation}
For a family $F\in\mathcal{F}$ with parameters $\theta$, we denote the shifted candidate by $x_{\mathit{shift}}=F(x;\theta)$, where $\theta$ bundles the relevant subsets of $(\theta_A,\theta_R,\theta_D)$ required by the chosen family.

To preserve visual plausibility, we enforce structural similarity (SSIM)~\cite{wang2004ssim} constraints both globally and within a foreground region of interest (ROI). Specifically, we derive a binary ROI mask $\Omega(x)\in\{0,1\}^{H\times W}$ from the input image using threshold-based foreground extraction~\cite{amiriebrahimabadi2024thresholding}, which separates the dominant anatomical foreground from background areas. For local patches $u$ and $v$, SSIM is defined as:
\begin{equation}
\mathrm{SSIM}(u,v)=
\frac{(2\mu_u\mu_v + C_1)(2\sigma_{uv}+C_2)}
{(\mu_u^2+\mu_v^2+C_1)(\sigma_u^2+\sigma_v^2+C_2)}
\label{eq:ssim}
\end{equation}
We compute $\mathrm{SSIM}(x,x')$ by averaging the SSIM map over all pixels, and $\mathrm{SSIM}_{\Omega}(x,x')$ by averaging only over pixels where $\Omega(x)=1$, providing a foreground-sensitive assessment that emphasizes clinically relevant structures:
\begin{equation}
\mathrm{SSIM}(x,x_{\mathit{shift}})\ge\tau,\ \mathrm{SSIM}_{\Omega}(x,x_{\mathit{shift}})\ge\tau
\label{eq:ssim_constraints}
\end{equation}
When $x_{\mathit{shift}}$ violates the constraints, we project it into the feasible set by mixing:
\begin{equation}
x_{\alpha}=(1-\alpha)\,x+\alpha\,x_{\mathit{shift}}\quad \alpha\in[0,1]
\label{eq:alpha_mix}
\end{equation}
Here $x_{\alpha}$ is the mixed sample and $\alpha$ controls the perturbation strength. We choose the largest feasible mixing coefficient via:
\begin{equation}
\alpha^{\ast}=\operatorname*{arg\,max}_{\alpha\in[0,1]}\ \alpha\quad
\text{s.t. }\mathrm{SSIM}(x,x_{\alpha})\ge\tau,\ \mathrm{SSIM}_{\Omega}(x,x_{\alpha})\ge\tau
\label{eq:alpha_star}
\end{equation}
and denote the final constrained sample by $x_{\mathit{adv}}=x_{\alpha^{\ast}}$.

\begin{figure*}[!t]
  \centering
  \includegraphics[width=\textwidth]{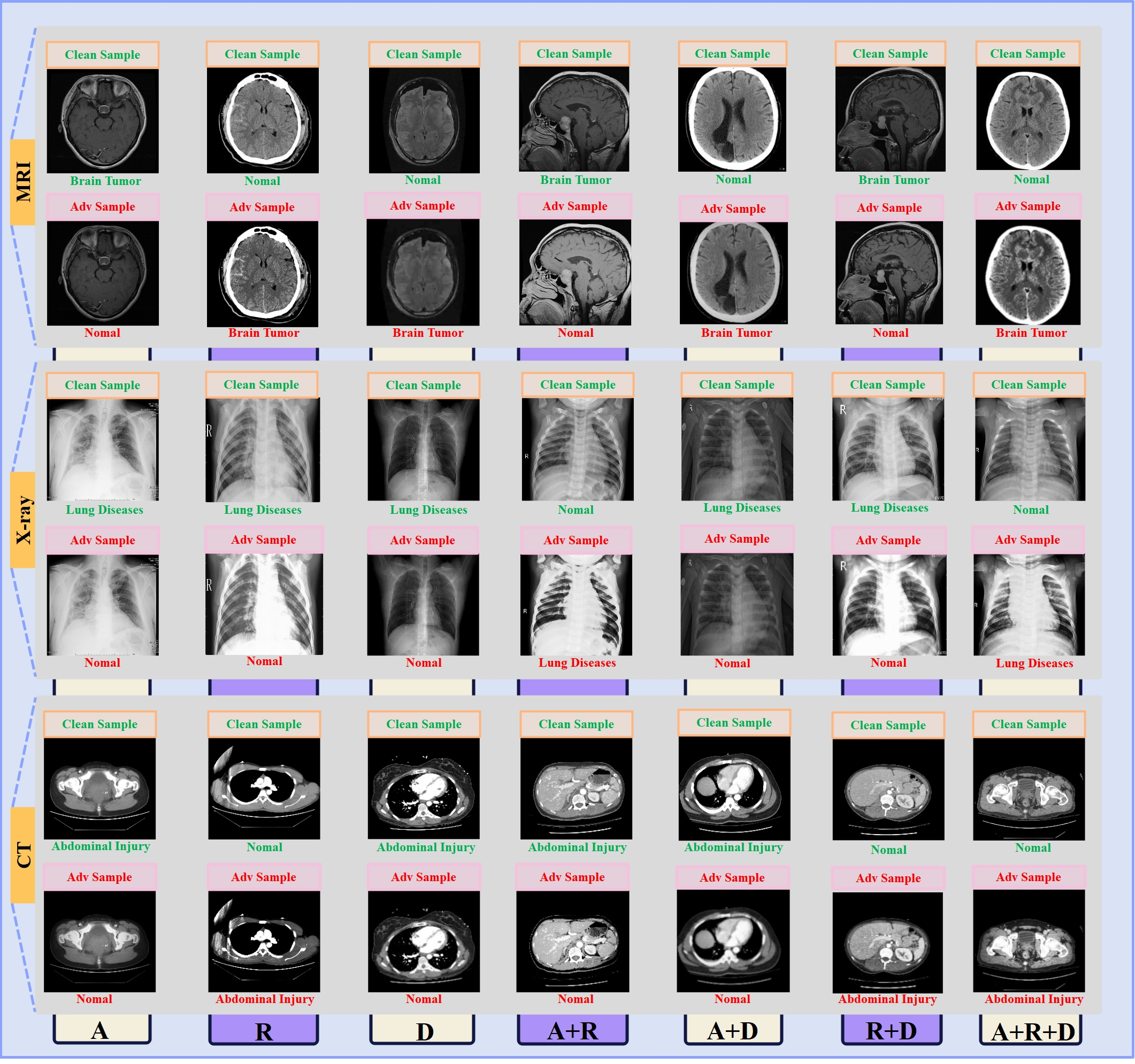}
  \caption{\textbf{Qualitative visualization of CoDA composition families.} Representative clean and adversarial samples for MRI, X-ray, and CT under the CoDA shift space in Eq.~\ref{eq:families_compact}.}
  \Description{A grid of examples for MRI, X-ray, and CT. Each column corresponds to one CoDA family. For each example, a clean sample (top) and its adversarial counterpart (bottom) are shown, along with the ground-truth label and the flipped prediction.}
  \label{fig:zs_vis_families}
\end{figure*}

\begin{table*}[!t]
  \caption{\textbf{Ablation under Bayesian optimization ($\tau=0.8$).} Zero-shot accuracy (\%) under single-stage shifts (Only $A$/Only $R$/Only $D$) versus the full chained CoDA shift ($A{+}R{+}D$).}
  \label{tab:ablation_single_stage}
  \centering
  \tblcommon
  \begin{tabular*}{\textwidth}{@{\extracolsep{\fill}}lccc ccc ccc ccc@{}}
    \toprule
    \multirow{2}{*}{Attack} &
    \multicolumn{3}{c}{BioMed-CLIP} &
    \multicolumn{3}{c}{UniMed-CLIP} &
    \multicolumn{3}{c}{BMC-CLIP} &
    \multicolumn{3}{c}{Rad-CLIP} \\
    \cmidrule(lr){2-4}\cmidrule(lr){5-7}\cmidrule(lr){8-10}\cmidrule(lr){11-13}
     & MRI & X-ray & CT & MRI & X-ray & CT & MRI & X-ray & CT & MRI & X-ray & CT \\
    \midrule
    Only $A$ & 75.5 & 61.0 & 47.5 & 62.0 & 56.5 & 44.0 & 77.0 & 48.0 & 63.0 & 73.0 & 68.0 & 46.0 \\
    Only $R$ & 57.0 & 33.0 & 31.0 & 56.0 & 49.0 & 31.5 & 61.5 & 41.5 & 48.0 & 56.5 & 40.0 & 35.0 \\
    Only $D$ & 76.5 & 60.5 & 40.5 & 56.0 & 52.5 & 37.5 & 69.5 & 42.0 & 58.0 & 65.0 & 52.0 & 39.0 \\
    \best{CoDA} & \best{44.0} & \best{24.5} & \best{22.0} & \best{44.0} & \best{42.5} & \best{8.5} & \best{57.0} & \best{32.5} & \best{40.0} & \best{50.5} & \best{28.0} & \best{30.0} \\
    \bottomrule
  \end{tabular*}
\end{table*}

\subsection{Optimization and Family Selection}
Let $x_{\mathit{adv}}(x;F,\theta,\tau)$ denote the constrained output obtained by applying family $F$ with parameters $\theta$ to $x$, followed by $\alpha$-projection under threshold $\tau$. For each image--label pair $\paren{x,y}$, family $F\in\mathcal{F}$, and threshold $\tau$, we optimize the within-family parameters to minimize signed correctness after projection, as illustrated in Fig.~\ref{fig:overview} (d):
\begin{equation}
\theta^{(F)}=\operatorname*{arg\,min}_{\theta\in\mathcal{S}_F}
J\bigl(x_{\mathit{adv}}(x;F,\theta,\tau),y\bigr)
\label{eq:theta_star}
\end{equation}
where $\mathcal{S}_F$ denotes the admissible parameter domain for family $F$. This formulation allows each family to be optimized within its own clinically plausible parameter range while treating the underlying model as a black box. It also decouples optimization over composition families from that over stage-specific parameters, enabling CoDA to capture both structural and parametric failure sources in realistic imaging workflows, where small deviations across stages may accumulate and interact nonlinearly. This separation allows CoDA to distinguish failures driven by stage composition from those caused by stage-wise parameter sensitivity within the same family. We optimize the objective using random optimization and TPE-based Bayesian optimization in Optuna~\cite{akiba2019optuna}.

After obtaining $\theta^{(F)}$ for each family, we select the final adversarial output from the family yielding the lowest signed correctness across all candidates:
\begin{equation}
F^{\ast}=\operatorname*{arg\,min}_{F\in\mathcal{F}}
J\bigl(x_{\mathit{adv}}(x;F,\theta^{(F)},\tau),y\bigr)
\label{eq:winner_family}
\end{equation}

\subsection{Post-hoc Repair via Teacher-Guided Token-Space Adaptation}
To mitigate failures without full retraining, we adapt a frozen student encoder with a lightweight residual token-space adapter, drawing on ideas from parameter-efficient adaptation methods such as LoRA and SSF~\cite{hu2022lora,lian2022ssf}, as well as knowledge distillation~\cite{gou2021kdsurvey} and efficient visual adaptation~\cite{he2023sensitivityaware,mercea2024timememory}. 

The overall repair workflow is summarized in Fig.~\ref{fig:token_space_repair}(a). Let $T_S(x)\in\mathbb{R}^{(1+N)\times d}$ denote the student's post-layernorm token sequence, where token index $0$ is the CLS token and the remaining $N$ tokens correspond to image patches. We insert a token-wise residual linear adapter:
\begin{equation}
\widetilde{T}_S(x) = T_S(x) + T_S(x) W^{\top}, \, W \in \mathbb{R}^{d \times d}
\label{eq:rotW}
\end{equation}
Here $W$ is the only trainable parameter, while the student backbone remains frozen throughout.

Let $P_S(x)\in\mathbb{R}^{N\times d}$ denote the patch tokens extracted from $T_S(x)$ after excluding the CLS token, and let $\widetilde{P}_S(x)$ denote the corresponding patch tokens extracted from $\widetilde{T}_S(x)$ for alignment. A teacher encoder provides patch tokens $P_T(x)$; when the teacher and student token dimensions differ, we apply a fixed projection to obtain projected teacher tokens $P'_T(x)$ in the student token space.

We define a token-geometry operator through the normalized Gram matrix of patch tokens:
\begin{equation}
\mathcal{G}(P)=\Normalize(P)\Normalize(P)^{\top}\in\mathbb{R}^{N\times N}
\label{eq:gram}
\end{equation}
where $\Normalize(\cdot)$ row-normalizes tokens to unit $\ell_2$ norm.

We train $W$ on clean images with three components: (i) a task loss $\mathcal{L}_{\mathit{task}}$ that preserves the original downstream objective, (ii) a view-consistency term weighted by $\lambda_{\mathit{cons}}$ that encourages invariant margins across two mild workflow-inspired augmentations, and (iii) an optional teacher-guided geometry alignment weighted by $\lambda_{\mathit{dist}}$ that matches the student and teacher patch-token similarity structure. The total objective is:
\begin{equation}
\mathcal{L}=\mathcal{L}_{\mathit{task}}
+\lambda_{\mathit{cons}}\bigl(m(x^{(1)})-m(x^{(2)})\bigr)^2
+\lambda_{\mathit{dist}}\bigl\|
\mathcal{G}(\widetilde{P}_S(x))-\mathcal{G}(P'_T(x))
\bigr\|_{F}^{2}
\label{eq:Ltotal_compact}
\end{equation}
Here $\|\cdot\|_{F}$ denotes the Frobenius norm. We evaluate repair directly on archived CoDA outputs without regenerating attacks.

\begin{figure*}[t]
  \centering
  \includegraphics[width=0.77\textwidth]{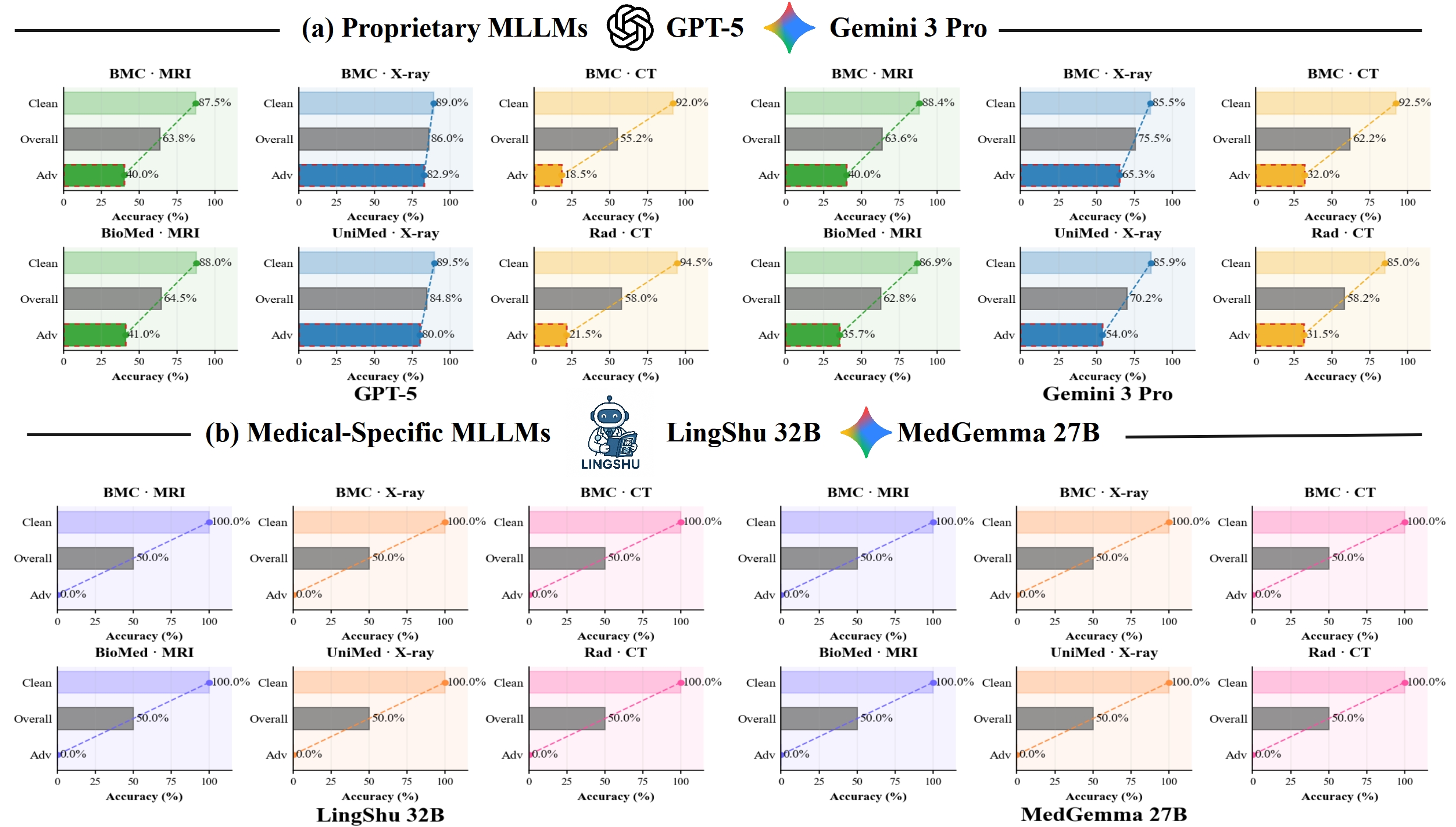}
  \caption{\textbf{Auditing performance under clean vs.\ CoDA shifts.} We compare proprietary and medical-specific MLLMs across MRI, X-ray, and CT, reporting clean accuracy, overall accuracy, and accuracy on CoDA-shifted inputs.}
  \Description{A grid of horizontal bar charts for multiple models and modalities. Each subplot reports clean, overall, and adversarial accuracy, comparing proprietary multimodal models and medical-specific multimodal models under CoDA shifts.}
  \label{fig:auditing_accuracy_grid}
\end{figure*}

% ============================================================
% 4 EXPERIMENTS
% ============================================================
\section{Experiments}

\subsection{Experimental Settings}

\textbf{Datasets.} We evaluate three imaging modalities using publicly available datasets spanning distinct anatomical regions. For brain MRI and chest X-ray, we follow the data organization protocol of Cheng et al.~\cite{cheng2025artefacts}, which curates public source collections for robustness evaluation. For abdominal CT, we use the RSNA Abdominal Traumatic Injury CT (RATIC) dataset~\cite{rudie2024ratic}. For each modality, we construct a balanced clean evaluation set with 100 positive and 100 negative images to more clearly isolate robustness effects under distribution shift~\cite{varma2024ravl,huang2025framesvqa,tu2025unicorn}.

\noindent\textbf{Threat models.} We evaluate CoDA under two plausibility thresholds, $\tau\in\{0.90,0.80\}$, across the seven composition families in Eq.~\ref{eq:families_compact}. For each image, the final adversarial output is obtained through constrained optimization followed by winner-family selection (Eq.~\ref{eq:winner_family}). The evaluation covers three complementary target settings. Zero-shot robustness is assessed on a diverse set of medical vision--language models, including BioMed-CLIP~\cite{zhang2023biomedclip}, UniMed-CLIP~\cite{khattak2024unimedclip}, BMC-CLIP~\cite{biomedica2025}, and Rad-CLIP~\cite{lu2024radclip}. Technical-authenticity auditing is conducted on proprietary MLLMs from the OpenAI GPT family~\cite{openai2023gpt4} and the Google Gemini family~\cite{gemini2023,gemini15_2024}, alongside medical-specific MLLMs including Lingshu~\cite{lingshu2025} and MedGemma~\cite{medgemma2025}. For post-hoc robustness repair, we select modality-specific teacher--student pairs based on attacked performance in the zero-shot setting and train a token-space adapter (Eq.~\ref{eq:rotW}) using the objective in Eq.~\ref{eq:Ltotal_compact}. Specifically, BMC-CLIP serves as the teacher for MRI and CT, whereas UniMed-CLIP serves as the teacher for X-ray; the student models are Rad-CLIP for MRI and X-ray, and UniMed-CLIP for CT. Repair performance is measured directly on archived CoDA outputs without regenerating adversarial samples at inference.

{\looseness=-1\noindent\textbf{Baseline methods.} For classification, we report clean accuracy and performance under CoDA at two plausibility thresholds, $\tau\in\{0.90,0.80\}$. At each $\tau$, we optimize CoDA using both random optimization and Bayesian optimization, and report the resulting zero-shot accuracy. In technical-authenticity auditing, we evaluate all auditors on clean inputs and on CoDA-shifted images generated by Bayesian optimization at $\tau=0.80$. For post-hoc repair, we compare a token-space adapter trained without teacher guidance against a teacher-guided variant with patch-token geometry alignment~\cite{gou2021kdsurvey,he2023sensitivityaware,mercea2024timememory} to improve robustness \mbox{in deployment}.}

{\looseness=-1\noindent\textbf{Implementation.} Constrained optimization is performed using random optimization and TPE-based Bayesian optimization in Optuna~\cite{akiba2019optuna}. For zero-shot classification, we evaluate brain MRI, chest X-ray, and abdominal CT in a binary disease-presence setting, using four semantically similar prompts for the positive and negative classes of each modality to reduce prompt sensitivity; prompts and pseudocode are given in the appendix. For technical-authenticity auditing, we adopt a single-call self-challenge protocol inspired by consistency-based prompting~\cite{wei2022cot,wang2022selfconsistency}, and report accuracy and reliability diagnostics; prompts appear in the appendix. For post-hoc repair, the adapter is trained only on clean images and tested on archived CoDA outputs without regenerating attacks, with $\lambda_{\text{dist}}=1$ and $\lambda_{\text{cons}}=0.1$ by default; pseudocode is provided in the appendix.}

\begin{figure*}[!t]
  \centering
  \includegraphics[
    width=\linewidth
  ]{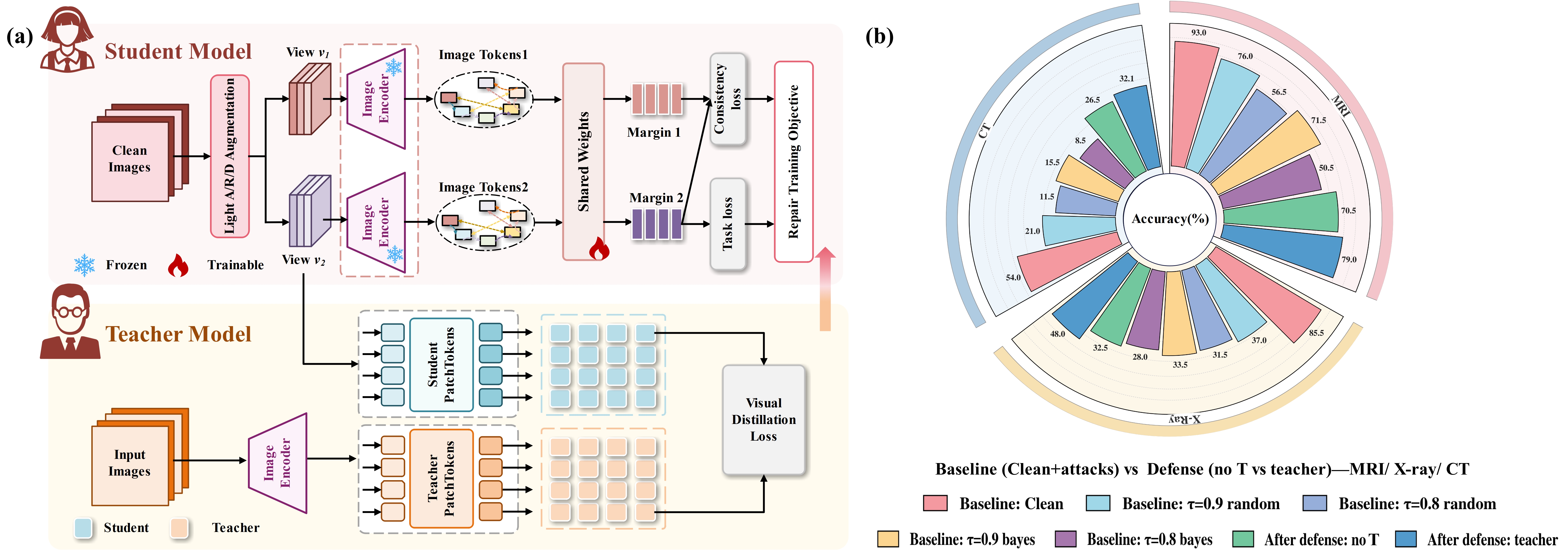}
  \caption{\textbf{Post-hoc token-space repair.} (a) A frozen teacher optionally guides a lightweight student token adapter trained on clean data with task, view-consistency, and distillation losses. (b) Accuracy on archived CoDA outputs for the baseline and repaired models, comparing unguided (no T) vs.\ teacher-guided (teacher) adaptation across MRI, X-ray, and CT.}
  \Description{A two-panel figure. Panel a shows a teacher-student diagram where the student uses a trainable token-space adapter and optionally matches teacher patch-token geometry. Panel b shows grouped bar charts for MRI, CT, and X-ray comparing baseline clean/attack performance to repair variants without teacher guidance and with teacher guidance.}
  \label{fig:token_space_repair}
\end{figure*}

\subsection{Zero-shot Classification Evaluation}

We evaluate zero-shot classification on both clean inputs and CoDA-shifted inputs across MRI, X-ray, and CT. Table~\ref{tab:family_success} summarizes the attack success rate of each composition family after optimizing the within-family CoDA parameters for every modality--model combination. CoDA achieves high success rates across all three modalities, with X-ray appearing the most vulnerable. For several models, composition families involving distortion nearly reach, or even achieve, perfect attack success. MRI and CT show a similar overall pattern, although the most effective family varies across models. This observation suggests that CoDA can induce failures through multiple plausible pipeline variants rather than through a single dominant transformation. Table~\ref{tab:zscls} reports Top-1 accuracy (\%) on clean inputs and under different CoDA settings, including random optimization and Bayesian optimization with $\tau \in \{0.90, 0.80\}$. Across all models and modalities, CoDA substantially reduces accuracy relative to the clean baseline. In most cases, Bayesian optimization is more effective than random optimization. These findings indicate that clinically plausible pipeline shifts are sufficient to consistently impair zero-shot recognition in CLIP-style MVLMs, with the largest degradation often observed on X-ray and CT. Figure~\ref{fig:zs_vis_families} presents representative clean and adversarial examples generated by the seven composition families in Eq.~\ref{eq:families_compact}. Although these perturbations largely preserve clinical readability, they introduce systematic appearance changes that are sufficient to alter, and in some cases even reverse, zero-shot predictions across modalities.

\subsection{Technical Authenticity Auditing Evaluation}
We evaluate MLLMs as technical-authenticity auditors. This non-clinical task asks whether an input image is a plausible product of clinical acquisition and delivery, without assessing pathology. We study both proprietary multimodal model families~\cite{openai2023gpt4,gemini2023,gemini15_2024} and medical-specific MLLMs~\cite{lingshu2025,medgemma2025}. We use a single-call protocol that emulates three interaction stages: an initial judgment with confidence, one sentence of technical evidence, and a symmetric peer challenge before the final decision, as shown in Fig.~\ref{fig:overview} (e-1). Beyond accuracy, we assess reliability using three indicators in Fig.~\ref{fig:overview} (e-2): contradiction rate, high-confidence error, and stubborn-wrong behavior. Results show a clear split between model groups. Proprietary MLLMs remain relatively reliable on clean inputs, but their performance degrades markedly on CoDA-shifted samples, with persistent high-confidence errors despite the built-in self-challenge step. In contrast, the medical-specific MLLMs exhibit severe deficiencies in technical-authenticity auditing. As shown in Fig.~\ref{fig:auditing_accuracy_grid}, their accuracy falls to 0 on CoDA-shifted MRI, X-ray, and CT images, indicating a complete failure to judge whether such inputs remain technically plausible for real-world clinical acquisition and deployment. Taken together, these findings suggest that current medical-domain specialization alone is insufficient for reliable authenticity auditing under distribution shift.

\subsection{Post-hoc Token-Space Repair Evaluation}
We evaluate whether CoDA-induced failures can be mitigated without full retraining. For each modality, we adapt the student encoder using the residual token-space adapter in Eq.~\ref{eq:rotW}, optionally enhanced with a stronger teacher through patch-token geometry alignment. Training uses only clean images, while evaluation is performed directly on archived CoDA outputs to reflect practical deployment constraints. Fig.~\ref{fig:token_space_repair} (b) compares the baseline student with the repaired variants on MRI, CT, and X-ray. The results show that token-space repair consistently improves robustness on adversarial inputs across all modalities, and that teacher-guided repair yields larger gains than the student-only variant. Nevertheless, a clear gap to clean performance remains, especially on X-ray and CT, indicating that lightweight post-hoc adaptation may not always efficiently recover from CoDA-induced failures.
% ============================================================
% 4.5 ABLATION STUDY
% ============================================================
\subsection{Ablation Study}
\label{sec:ablation}

% Avoid overfull lines locally (prevents text spilling into the next column)
{\setlength{\emergencystretch}{1.2em}

\noindent\textbf{Ablation of Optimization Iterations.}
We vary the iteration budget used by the constrained optimization routine and quantify how attack effectiveness scales with additional optimization steps.
Fig.~\ref{fig:ablation_iters} reports the attack success rate as a function of iteration count for MRI, X-ray, and CT across four CLIP-style MVLMs.
Increasing the budget consistently improves success, while gains taper after moderate iteration counts, indicating that most improvements can be achieved within the masked-SSIM feasible set.

\begin{figure}[!t]
  \centering
  \includegraphics[width=\linewidth]{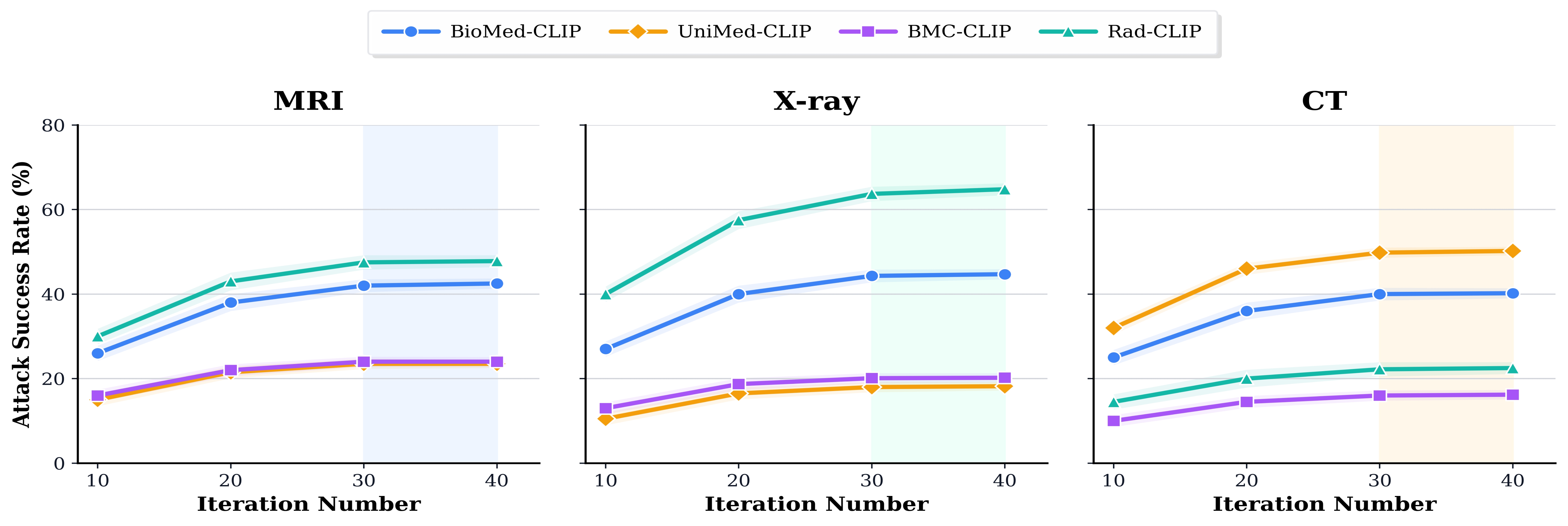}
  \caption{\textbf{Ablation of optimization iterations.} Attack success rate under CoDA as a function of the optimization iteration budget, reported for MRI, X-ray, and CT across BioMed-CLIP, UniMed-CLIP, BMC-CLIP, and Rad-CLIP.}
  \Description{A three-panel line chart for MRI, X-ray, and CT showing attack success rate versus iteration number, with one curve per model.}
  \label{fig:ablation_iters}
\end{figure}

\noindent\textbf{Ablation of Single-Stage Shifts vs.\ Full CoDA ($\tau=0.8$).}
To isolate the contribution of each pipeline component, we restrict the shift space to a single stage (Only $A$, Only $R$, or Only $D$) and compare against the full chained composition under the same plausibility threshold.
Table~\ref{tab:ablation_single_stage} reports the resulting accuracies.
Across models and modalities, the chained composition remains the most damaging setting, supporting the hypothesis that cross-stage interactions amplify brittleness beyond what any single stage induces.

\noindent\textbf{Ablation of Repair Hyperparameters.} We ablate two key hyperparameters in post-hoc token-space repair: the view-consistency weight $\lambda_{\mathit{cons}}$ and the teacher-guidance weight $\lambda_{\mathit{dist}}$ in Eq.~\ref{eq:Ltotal_compact}. Fig.~\ref{fig:defense_ablation} shows that, across modalities, post-repair accuracy under CoDA-shifted inputs generally lies between the clean and attack baselines, exceeding the latter in most cases. In a few CT settings, the repaired model slightly outperforms the clean baseline, possibly because the student model itself shows relatively weak performance under clean baseline conditions in practice.

\begin{figure}[!t]
  \centering
  \includegraphics[width=\linewidth]{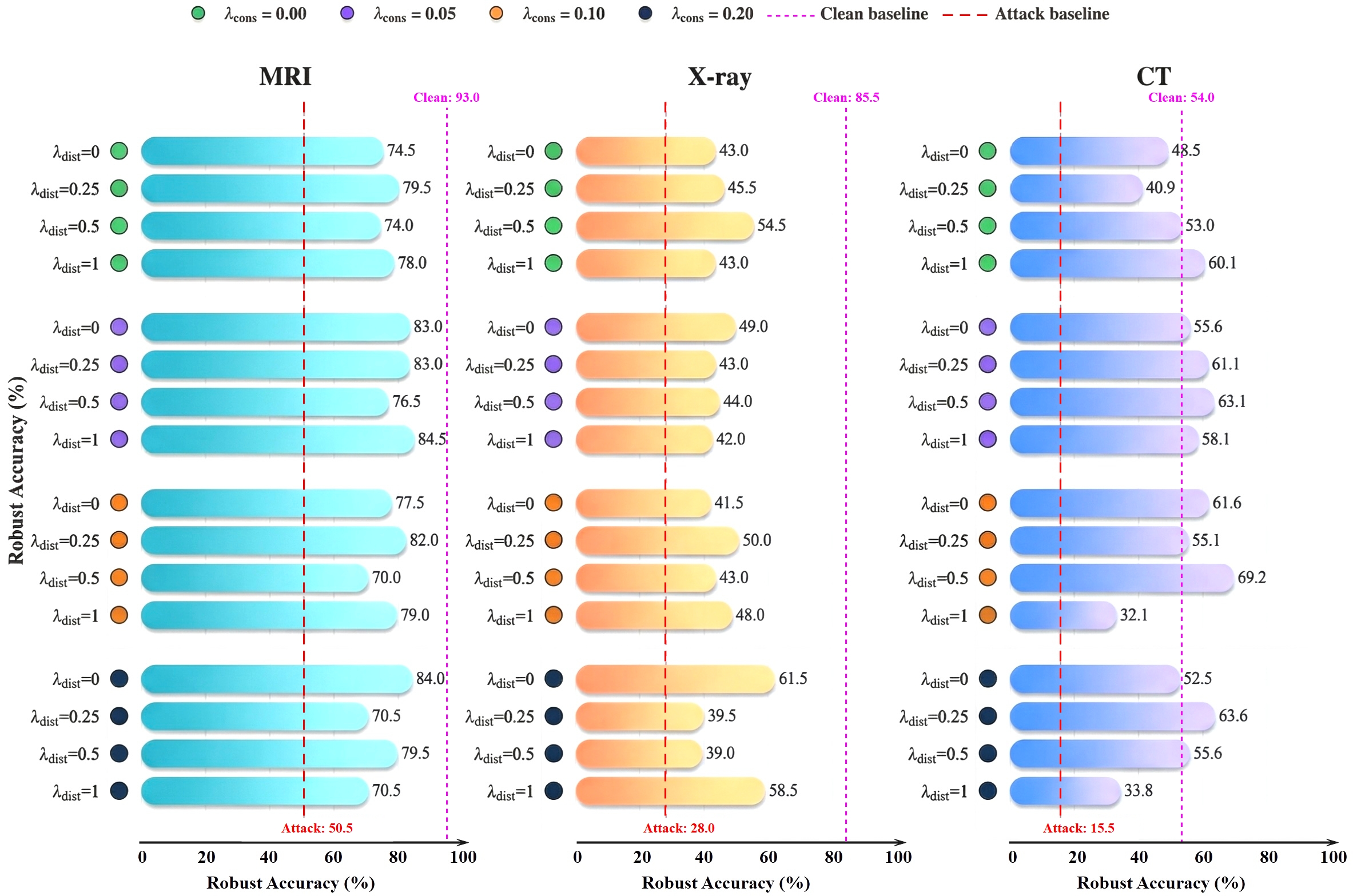}
  \caption{\textbf{Ablation of repair hyperparameters.} Robust accuracy under CoDA shifts as a function of $\lambda_{\mathit{cons}}$ and $\lambda_{\mathit{dist}}$, compared against the clean baseline and the attack baseline, reported for MRI/X-ray/CT.}
  \Description{A three-panel bar-chart figure for MRI, X-ray, and CT. Bars show robust accuracy under different settings of the view-consistency weight and teacher-guidance weight, with a clean baseline region and an attack baseline indicated by a red dashed line.}
  \label{fig:defense_ablation}
\end{figure}

} % end local emergencystretch group

% ============================================================
% 5 CONCLUSION
% ============================================================
\section{Conclusion}
We present CoDA, a chain-of-distribution framework that composes clinically plausible pipeline shifts and optimizes failure-inducing parameterizations under masked similarity constraints. Across MRI, CT, and X-ray, CoDA substantially degrades the zero-shot performance of CLIP-style MVLMs, revealing vulnerabilities often overlooked by standard single-stage evaluations. We further show that MLLMs, when used as technical-authenticity auditors, are also susceptible to such shifts, including persistent high-confidence errors. Our results also indicate that a lightweight post-hoc repair strategy based on teacher-guided token-space adaptation can partially recover robustness on archived CoDA adversarial outputs, offering practical support for more robust real-world deployment.
% ============================================================
% References
% ============================================================
\bibliographystyle{ACM-Reference-Format}
\bibliography{refs}

\end{document}